\documentclass[a4paper, 10pt, conference]{ieeeconf}      

\IEEEoverridecommandlockouts                              

\usepackage{graphicx}
\usepackage[export]{adjustbox}

\usepackage{url}

\title{\LARGE \bf
Teat Pose Estimation via RGBD Segmentation for Automated Milking
}

\author{Nicolas Borla$^{1}$, Fabian Kuster$^{1}$, Jonas Langenegger$^{1}$, Juan Ribera$^{2}$, Marcel Honegger$^{1}$, Giovanni Toffetti$^{2}$ \\

\thanks{This research project is supported by Innosuisse, the Swiss Innovation Agency, with project number 40368.1 IP-ENG.}
\thanks{$^{1}$ The authors are with the Institute of Mechatronic Systems (IMS),  Technikumstrasse 5,
8401 Winterthur,  Switzerland}
\thanks{$^{2}$ The authors are with the Institute of Applied Information Technology (InIT),  Obere   Kirchgasse   2,   8400 Winterthur,  Switzerland}

Zurich University of Applied Sciences (ZHAW)
}

\begin{document}
\maketitle
\thispagestyle{empty}
\pagestyle{empty}

\begin{abstract}

We present initial results in the development of a novel robot using RGBD cameras, image segmentation, and a simple teat pose estimation algorithm for automated milking.
We relate on the analysis of the accuracy of different commercial RGBD cameras in realistic conditions.
Although preliminary, our initial implementation shows that 2D image segmentation combined with point cloud processing can achieve repeatable millimeter-scale precision in estimating (synthetic) teat tip positions and cup attachment approach.
The solution is also applicable in a cloud robotics setup, with GPU-based segmentation executed on an edge device or cloud.

\end{abstract}

\section{Introduction}\label{sec:introduction}

Milking robots have been in use for almost 30 years, after the first systems were installed in the Netherlands in 1992 \cite{bijl2007dairyscience}. Today, the main suppliers of milking robots are Lely, DeLaval and GEA Farm Technologies. Their systems are typically used on large dairy farms and are optimized for milking cows up to 3 times per day, and around the clock.
However, in many European countries such as Switzerland, dairy farms typically have less than 60 cows and they are usually milked only twice, once in the morning and once in the evening, while they are left to graze during the day. This use case requires milking robots that occupy less space, so that it is easier to install several systems in parallel in existing barns, and it requires the robots to milk cows as efficiently as possible, because the time slots for milking are far shorter. For this reason, we started a research project to design a new generation of milking robot, using the latest technologies to reduce the space required for the milking robot manipulator, and to reduce the time required to milk cows.\\
While most existing milking robots only offer 3 degrees of freedom (DoF) \cite{kuhlbusch2006fluidpower}, this new manipulator offers 5 DoF, so that the milking cups can be positioned in all Cartesian coordinates, and their orientation aligned to the orientation of the teats. This allows to more reliably place and attach the milking cups to the teats.\\
\begin{figure}[h!]
    \centering
    \includegraphics[width=0.7\linewidth, trim={0.1cm 0.1cm 0.1cm 0.1cm}]{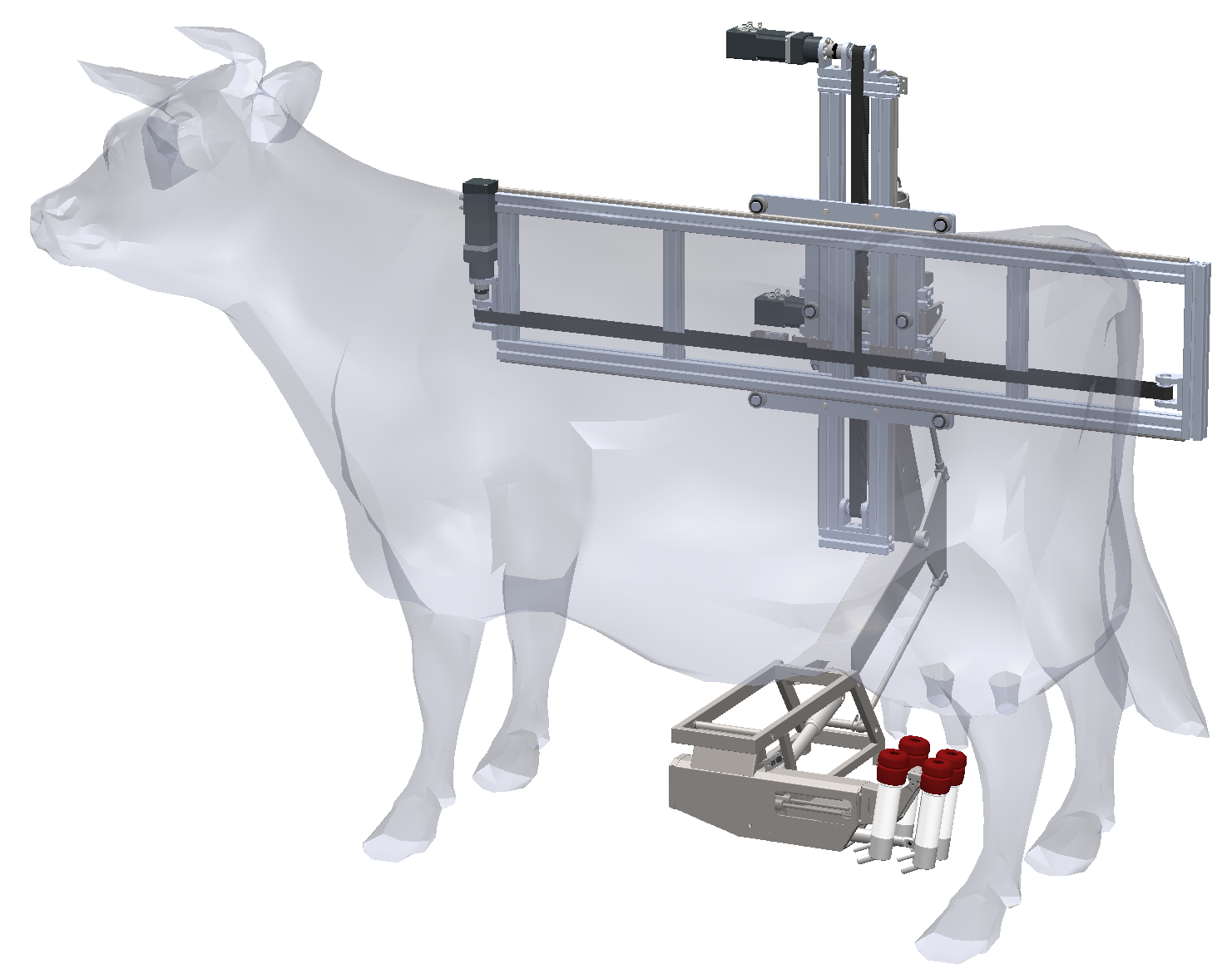}
    \caption{Kinematic model of the milking robot}
    \label{fig:robot}
\end{figure}
Attaching milking cups to teats often takes more than a minute with existing milking robots \cite{hugle1999land}. Before attaching cups, these robots often scan the teats with laser scanners to detect the teats positions, then the manipulator moves the cups to the teats for attachment. Attachments fail in 5 to 10\% of all cases \cite{hugle1999land}, which requires another scan of the teats, for another trial to attach the cups.\\
With a new approach to detect the teat poses with RGBD cameras in real-time, attaching the cups to teats does not require a preceding, time-consuming  scanning procedure. Instead, the teat positions and orientations are detected while the robot manipulator is moving the cups to the estimated position of the teats, and is adjusting the motion to the detected positions with each measurement. This will reduce the time needed to attach the cups to the teats, and increase the reliability of the attachment process significantly.\\

\section{Related Work}\label{sec:relatedwork}



In the following subsections we relate on the state of the art of commercial automated milking robots, published research in the field, and more general algorithmic approaches that can be applied to the problem at hand.

\subsection{Commercial Milking Robots}

Five major international commercial milking robot suppliers provide their services worldwide:
Lely, DeLaval, GEA, SAC, and Lemmer-Fullwood.
Apart from them, there are several smaller (typically national) suppliers. A complete discussion of the pros and cons of each solution is beyond the scope of this paper, moreover there are several specialized publications offering such comparisons online\footnote{E.g., \url{https://www.melkroboter.net} (in German)}.

Overall, the major drawbacks common to all current commercial solutions are:

\begin{itemize}
    \item High cost per milking robot unit;
    \item Intense robot usage to amortize investment cost;
    \item High cost of replacement parts and materials;
    \item Limited or no support for continuous learning / adaptation to cow udder morphology (i.e., personalization)
\end{itemize}

\subsection{Automated Milking Literature}

Many of the current state of the art solutions rely on laser scanner technology to detect teats and estimate their pose. A 2D laser scanner implies a scanning procedure, moving the sensor to different heights to achieve a 3D measurement. A significant drawback of this design is that the measurement cannot be performed in real-time. If the cow moves, a new measurement procedure is needed before the cups can be attached.
Relying purely on depth information, laser technology may fail in correct teat identification, therefore manipulating the suction cups in the wrong direction. For this reason, \cite{pal2017algorithm} proposes a fast and reliable solution to the problem using Time of Flight (ToF), RGBD and Thermal Imaging. 
The study from \cite{akhloufi20143d} for vision systems for livestock reports that RGB-D technologies are preferable to ToF cameras.

Similarly, \cite{rastogi2019teat} takes a stand against the limitations of laser assisted edge detection technologies, which cannot differentiate between a healthy and a diseased teat. They propose two alternatives to the task: a Haar-cascade classifier and a YOLO classifier for cow teats. Both approaches work on real time but lack reliable accuracy. 

In \cite{vanderzwan2015} several references are given to teat pose estimation algorithms applied by commercial milking robots. However, the authors state that their method to identify teat tip positions from low resolution 3D-ToF camera videos is superior to all previously reported ones. The method is based on edge detection on the depth image combined with matching U-shaped templates. To account for teat size and distance from camera, resized U-shapes are applied  for correlation. In order to account for non-vertical teats, PCA (principal component analysis) is used to obtain rotational invariant teats. The proposed solution requires limited computation and achieves teat pose estimation at 4 to 8 FPS. Validation results show ``90\% of the frames being successfully tracked'' on 15 videos.

The work in \cite{OMahony2019}, discusses the application of 3D vision technologies to precision livestock farming, including in automated milking, and concludes that at time of publication (2019) 3D deep learning solutions were not yet applicable due to a lack of sufficient training data, a problem common to all 3D deep learning computer vision applications.

Finally, in \cite{Dorokhov2020}, the authors use a 3D-ToF camera 
to collect both RGB images and a point cloud. They process the point cloud applying the k-nearest algorithm for segmentation, but such method cannot distinguish the udder from the teats, resulting in imprecise segmentation. To counter this problem, their method relies on assumptions on the camera position w.r.t. the teat for teat detection. Still, this prevents them from correctly identifying teats on real cows where udder morphology is highly variant. 

\subsection{Object Detection, Pose Estimation, Grasping}

As reported in \cite{singh2018fotonnet}, "the ability for robots and computers to see and understand the environment is becoming a burgeoning field, needed in autonomous vehicles, augmented reality, drones, facial recognition, and robotic helpers". Since the rise of the CNN \cite{krizhevsky2017imagenet} deep learning based methods for image classification have reached state of the art performance for 2D detection. Nevertheless, 3D scene interpretation methods continue to struggle because of 1) the lack of publicly available RGB-D data sets \cite{han2019image} and 2) the not yet widespread adoption of depth cameras compared to 2D ones \cite{singh2018fotonnet}.

Several algorithms have been developed for automated pose estimation and / or grasp generation of objects in literature, for instance \cite{ciocarlie2014towards}, \cite{mahler2017dex}, and \cite{gpd}.
The approaches above address the general problem of grasping and manipulating unknown objects, however they cannot directly be applied to our specific manipulation task (i.e., teat attachment and successful pumping).
One possible approach that goes in the direction of generalizing object classes and their manipulation is \cite{manuelli2019kpam} 
which uses semantic 3D keypoints for object representation and enables the specification for robot action planning and grasping with centimeter level precision. Our initial work done in this paper is a needed step to try to apply that kind of approach to automated milking.

\section{Requirements}\label{sec:reqs}

The initial requirements from the project specification for the teat pose estimation algorithm prototype were:
\begin{itemize}
    \item Pose estimation must be performed continuously during the movement of the robotic arm;
    \item Maximum estimation time of all teat poses of 10 seconds (this includes any arm movement required to reduce uncertainty);
    \item Recognition of the correct pose (within an error of 0.5 centimeters) in at least 90\% of teats.
\end{itemize}

Apart from these formal requirements coming from the project contract, further requirements for the solution stemmed from the fact that the robot shape and kinematics had to be designed, hence further requirements for the algorithm are:

\begin{itemize}
    \item No assumptions should be made about camera(s) positions and orientations w.r.t. the udder;
    \item Occlusions have to be expected and taken into account;
    \item No assumptions should be made about teat number\footnote{Not all cows have exactly 4 teats \cite{Dorokhov2020}}, positions, and orientations (cow's udder morphology);
\end{itemize}

No initial requirements were given with respect to the architecture and cost of the compute unit (or GPU) to be used for the implementation.
\section{Solution Architecture}\label{sec:design}

Given the requirements from the previous section, and our analysis of the state of the art from Section \ref{sec:relatedwork}, we oriented ourselves in choosing a solution that would allow us to minimize assumptions (e.g., on poses / frames) and at the same time account for natural variation (i.e., changing udder morphology, teat colors, light conditions).
This lead us to restrict the space of solutions towards a combination of Neural Networks (NNs) based on Deep Learning (DL) with pose estimation from point clouds (PCLs).

In this respect, our review of the literature of DL solutions applied to 3D convinced us that, at that specific moment, we could not leverage any existing 3D DL technology for the project, be it for reasons of prediction performance (both in terms of rate and accuracy) or training cost (including training data set labelling).

Upon this reasoning, we decided to build on existing mature DL technologies to identify teats from 2D color images. The rationale here is that 2D DL allows us to minimize false positives in recognition while accounting for natural variation of morphology, colors, and light conditions. Here, discounting the different NN models, the main decision to make was whether to use multi-object identification (i.e., bounding boxes around teats) or multi-object segmentation (i.e., a pixel mask for all pixels belonging to each teat). We opted for the second alternative which, albeit slower (e.g., with Mask-RCNN), allowed us to have a more precise mask closely matching the shape of the teats as seen in 2D.

To bridge the gap from 2D to 3D, we borrowed the idea from \cite{Qi2018} to project the mask stemming from the 2D teat identification step into the point cloud with a frustum to ``carve out'' teats in three dimensions.
Then, for each 3D teat candidate, a combination of clustering, PCA, and surface normals algorithms can be used to estimate the orientation of a teat, identify the tip, and compute the its 6 DOF pose in 3D space.

\begin{figure}[ht]
    \centering
    \includegraphics[width=0.9\linewidth]{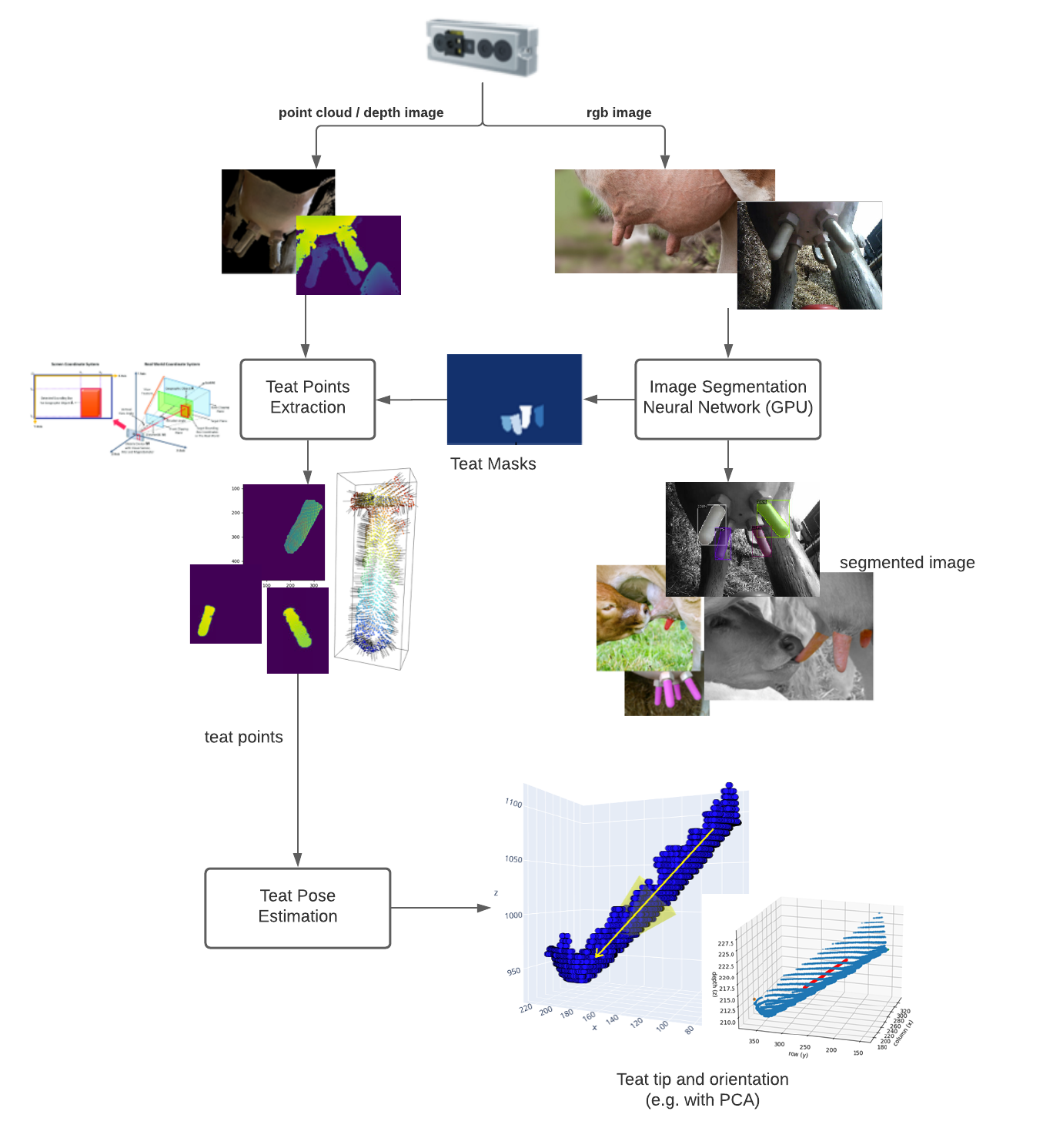}
    \caption{Main functional blocks of our teat pose estimation solution}
    \label{fig:design}
\end{figure}

The resulting high level functioning of our solution is depicted in Figure \ref{fig:design}.

\section{Implementation}\label{sec:implementation}


The implementation of the solution was distributed across the labs participating in the project based on expertise. ICCLab (InIT) focused on the teat pose estimation algorithm, while the IMS took responsibility of all the robotic aspects, from the evaluation of different cameras for the task at hand, to their calibration and correction of errors, to the design of the final robot and the programming of the arm control logic.

Given the distributed nature of the project and the possibility to apply cloud robotics solutions to the final product, we decided to implement the software stack for the project as a distributed system from the get go. In particular, we used containerization and a multi-master ROS design to isolate the different versions of operating systems, ROS, and libraries that were needed for the different components of the project.

\begin{figure}[ht]
    \includegraphics[width=1.0\linewidth]{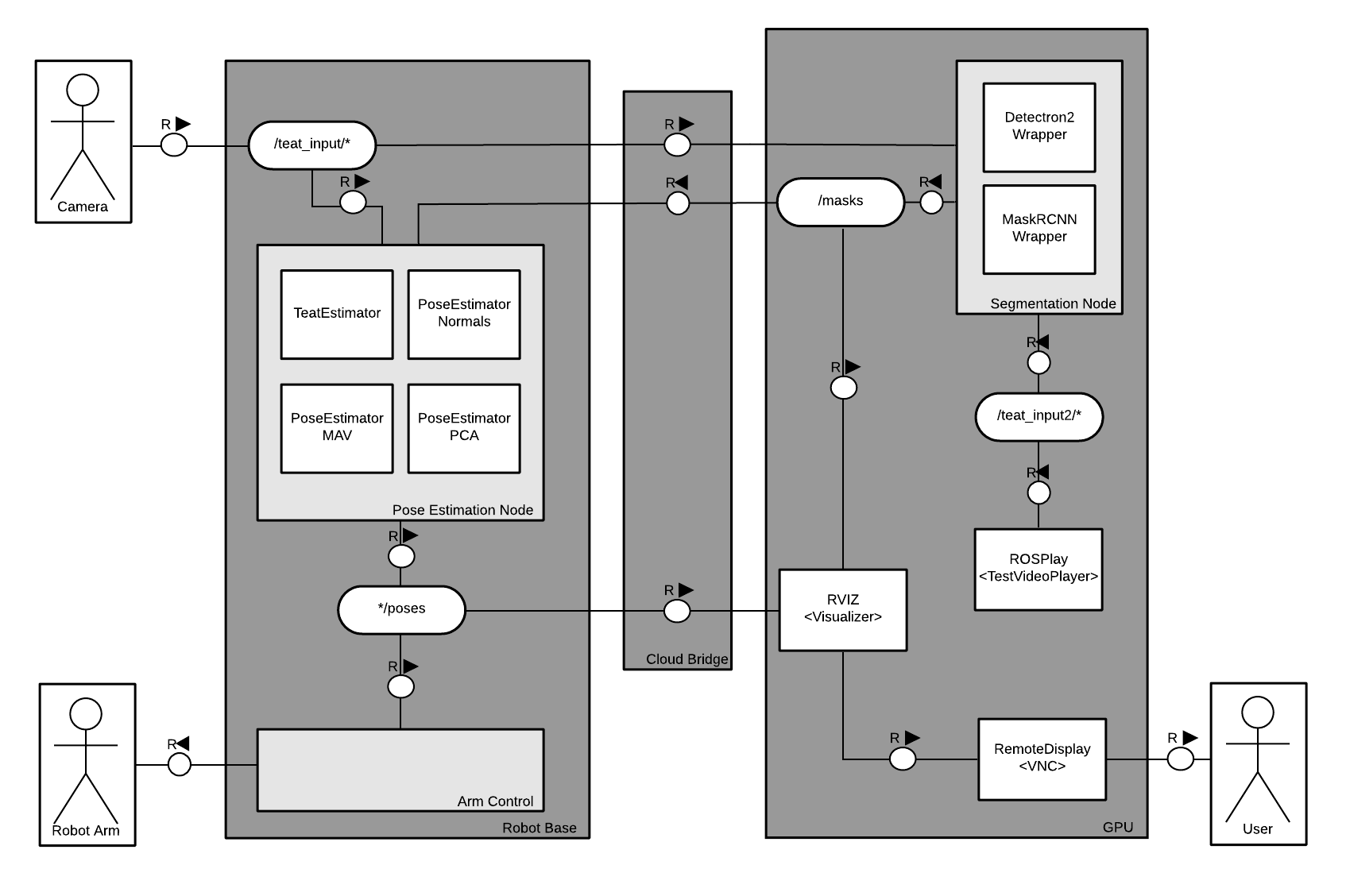}
    \caption{FMC architectural diagram of our teat pose estimation solution}
    \label{fig:arch}
\end{figure}

The overall component architecture of the final implementation is depicted in Figure \ref{fig:arch}.

We opted for using three different ROS nodes each implementing a subset of the functionalities: 
\begin{itemize}
    \item the \emph{Pose Estimation node} is the interface to the sensor data and the robot. It receives the (time synchronized) messages from the camera sensors (i.e., RGB image and point cloud), forwards the RBG image to the NN, awaits for the teat masks to be detected, and uses the masks to publish the detected teat poses;
    \item the \emph{Segmentation node} hosts the NN that performs the multi-object image segmentation and publishes the detected masks;
    \item the \emph{Arm Control node} receives multiple messages about estimated teat poses and performs arm movement planning if a configurable number of teat pose estimates is consistent over time.
\end{itemize}

In the following subsections we relate on the implementation of each of the nodes.

\subsection{Pose Estimation Node  (find\_teat\_poses)}

This is the node that connects the robot to the segmentation neural network node. A synchronized message filter receives both the point cloud and  the rgbImage published at the same time instant. The node saves the point cloud in memory and forwards the rgbImage to be processed by the NN.

After segmentation, the NN publishes the segmented image for visualization and the "masks" resulting from segmentation. We use a project specific ROS message format to reduce the amount of data that is passed back from the NN to represent the masks.

Upon reception of the masks the node find\_teat\_poses uses the 2D mask contour to extract the corresponding 3D points from the point cloud and to estimate the position and orientation of all visible teats.
Teat points are extracted from the point cloud by first applying a voxelization step, then projecting the rays matching each teat masks contour in 3D space and removing anything outside of the generated frustum.

\begin{figure}[ht]
    \centering
    \includegraphics[width=0.8\linewidth]{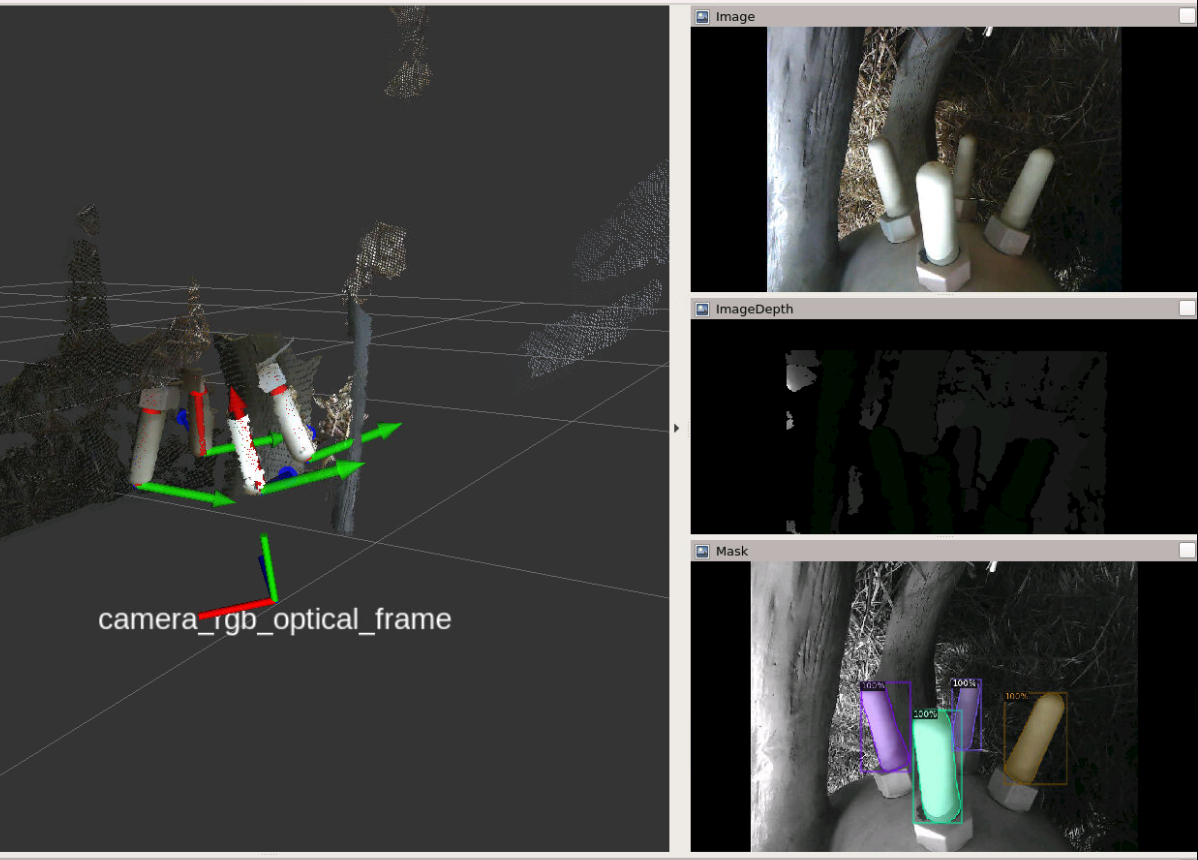}
    \caption{Visualization of teat pose markers and teat segmentation}
    \label{fig:markers}
\end{figure}

Finally a set of pose estimation algorithms are applied to estimate teat tip positions and the required orientation of the teat cup for a successful attachment.  All teats poses are published and visualized with a marker as in Figure \ref{fig:markers}.

\textbf{Teat Tip Pose Estimation Algorithms:}
With the 3D points for each teat as input, different algorithms can be applied to understand the location of the teat tip and the required orientation of a teat cup to perform an attachment. We relied on two simple implementations from geometric principles: principal component analysis (PCA) and using surface normals.

With the PCA algorithm, the underlying assumption is that cow teats have a generic cylindrical shape and are longer than wider, hence running PCA on the 3D points of a teat would yield the teat ``cylinder axis'' as the main principal component.
Given the fact that the teat axis can be represented by two vectors with opposite orientation, we use the camera position to identify the vector direction for cup attachment (i.e., upwards rather than downwards) hence the teat tip.

The surface normals algorithm is based on a different idea to identify a teat axis. That is, if a cow teat is approximately cylindrical, the vectors that are orthogonal to its surface (i.e., "surface normals") will also be normal to the teat axis. Hence, the teat axis direction can be estimated by finding the vector that minimizes the sum of the dot products with the vector itself for each surface normal.
As in the case of the PCA algorithm, a further step to correctly identify the vector orientation for teat attachment has to be performed.

\subsection{Segmentation Node (Neural Network)}

In the course of the project we experimented with different publicly available neural network implementations to perform either object detection or multi-object segmentation.

In the end we built our prototype based at on two implementations of the MaskRCNN paper: Matterport's MaskRCNN \footnote{\url{https://github.com/matterport/Mask_RCNN}} and Facebook Research's Detectron2 implementation of MaskRCNN\footnote{\url{https://github.com/facebookresearch/detectron2}}.

Both implementations are highly configurable and allowed us to use 640x480 pixel images as input wrapping the invocation of their inference functionality in a simple ROS topic subscriber callback handler.

Benchmarks\footnote{\url{https://github.com/facebookresearch/maskrcnn-benchmark/issues/449}} of both implementations show clear differences in the Average Precision (AP) between them 
showing Detectron2 having better accuracy. 
Moreover, benchmarks show the implementations from matterport have a 4x slower throughput (imgs/sec) compared to Detectron2\footnote{\url{https://detectron2.readthedocs.io/en/latest/notes/benchmarks.html}}. These drawbacks and the generally better performance of the Detectron2 implementation led to it being the favorite for the segmentation task.

\section{Results}\label{sec:evaluation} 

In this section we relate on the methodology and results obtained in evaluating different commercial cameras and implementing a first prototype of a complete teat pose estimation and attachment solution.

\subsection{3D sensor}

An essential requirement for the 3D sensor mentioned in the previous sections is that pose estimation must be performed on the fly during arm (and cow) movement. Therefore, no sensor which needs a scanning procedure is suitable. 
We selected five 3D cameras among the newest models available on the market and carried out an accurate evaluation of their performance to choose the most suitable for our task. These cameras are produced by different manufacturers and use various measurement technologies:
\begin{itemize}
    \item Orbbec Astra Embedded S
    \item Orbbec Astra Stereo S U3
    \item PMD/Infineon CamBoard Pico Flexx
    \item Intel RealSense SR305
    \item Intel RealSense D435
\end{itemize}

\begin{figure}[t]
    \includegraphics[width=1.0\linewidth]{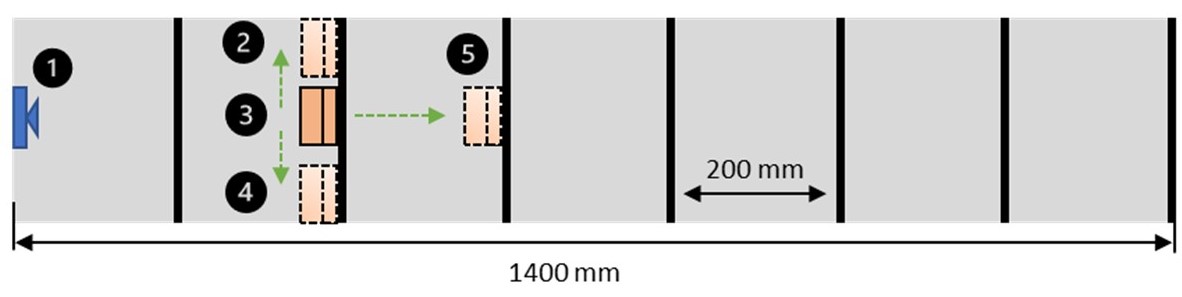}
    \caption{Test setup used to evaluate the performance of different cameras: (1) 3D camera, (2) test object placed in the left track, (3) test object placed in the central track, (4) test object placed in the right track, (5) the distance is measured placing the test object further away from the camera (every 200mm)}
    \label{fig:camera_test_setup}
\end{figure}

To estimate distances, the two cameras of Orbbec and the Intel D435 use active IR stereo vision, the Pico Flexx uses an IR Time-of-Flight sensor, and the Intel SR305 uses structured light. All cameras except the Pico Flexx integrates an RGB camera, meaning that they provide data in the form of 3D point clouds and 2D colour images.\\
To evaluate the five cameras, we used a test setup already available at IMS for the test of general purpose 3D sensors \cite{Cortesi2019}. This setup consists of a steel plate with plastic stops every 200mm up to a distance of 1.4m with three tracks: left, centre and right. The test object is a 3D printed L-form  with a surface of 100x150mm (the surface is orthogonal to the camera). The test object is placed every 200mm, and the distance is measured by averaging the points measured on the surface. Figure \ref{fig:camera_test_setup} illustrates the concept of the test setup.

The absolute accuracy of the distance measurement for all cameras is evaluated under different conditions using the test setup. The different situation evaluated are:
\begin{itemize}
    \item Different light condition (direct light from headlamp, room light or night)
    \item Different colour of the test object (White, Black or Pink like the teats of the cow)
    \item Lateral shift (change track)
    \item Influence of a 5mm glass panel in front of the camera (to simulate the sealing needed to work outdoor)
    \item Variance (repeatability of the measurement)
    \item Influence of the camera resolution (for the cameras which offer a configurable resolution)
\end{itemize}

\begin{figure}[ht]
    \includegraphics[width=1.0\linewidth]{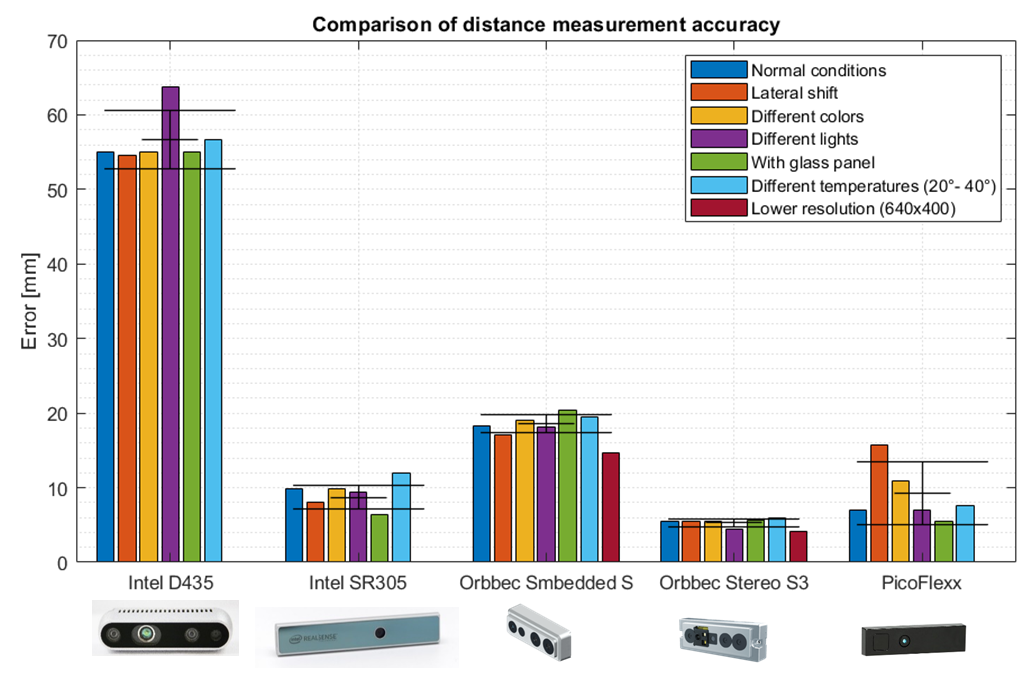}
    \caption{Maximal distance measurement error for objects up to 1m under diverse conditions}
    \label{fig:Accuracy_up_to_1_m}
\end{figure}

The results of the test are shown in figure \ref{fig:Accuracy_up_to_1_m}. In this chart, only the maximal error for distances up to 1m is considered for all different conditions and cameras. The reasons are that the robot's working range under the cow is limited to 1m in the mechanical requirements and that it makes it easier to show the accuracy of the cameras and the influence of the test conditions.\\
As a general behaviour, all cameras show an error in absolute accuracy that increases approximately quadratically with distance.
The maximum error at 1m distance, shown in figure \ref{fig:Accuracy_up_to_1_m}, also reflects how fast the error grows for each camera.\\
The variance is not illustrated because under steady condition is less than 0.2mm for all cameras. It is not the repeated measurement that introduces a relevant error in the measurement, but rather a change in the measurement conditions.\\
It can be noted that the PicoFlexx, the Intel SR305 and the Orbecc stereo have similar performances and overall are better than the other two cameras. PicoFlexx uses IR Time of Flight technology and seems to be more sensitive to different working conditions. Moreover, this camera has no RGB sensor included. The Intel SR305 is much larger than The Orbbec Stereo (at least twice the volume) and less accurate. Therefore, we chose the Orbbec stereo for our implementation. This camera performs overall better than all other sensors tested, has an RGB camera already incorporated, and according to the manufacturer is specifically developed to work in a multi-camera setup. A setup with multiple cameras could be interesting for the milking robot, and therefore, the same tests were repeated using two Orbbec stereo pointing at the same object. The results showed no interference between the cameras and the same result as the setup with only one camera.  

\subsection{Pose Estimation Accuracy}
In the current first phase of the project, experiments were conducted with a dummy cow under laboratory conditions. This framework implies an indoor environment, varying (low) light conditions, and no varying udder geometry. We used the UR10e robot as a manipulator with the Orbbec Astra Stereo S U3 camera and a single teat cup mounted on the robot flange. An Ubuntu computer was selected as a local controller with the ROS framework for software development. The 2D camera image is sent over wifi to a separate virtual machine in our local cloud computing cluster equipped with a Nvidia Tesla T4 graphic card for teat detection. Upon detection, teat masks are used to predict teat poses on the robot. To validate the four calculated teat positions, we moved the robot with the attached teat cup to each teat of the udder one after the other. \\

\begin{figure}[h!]
    \includegraphics[width=1.0\linewidth, trim={1cm 10.9cm 1.2cm 11.3cm}]{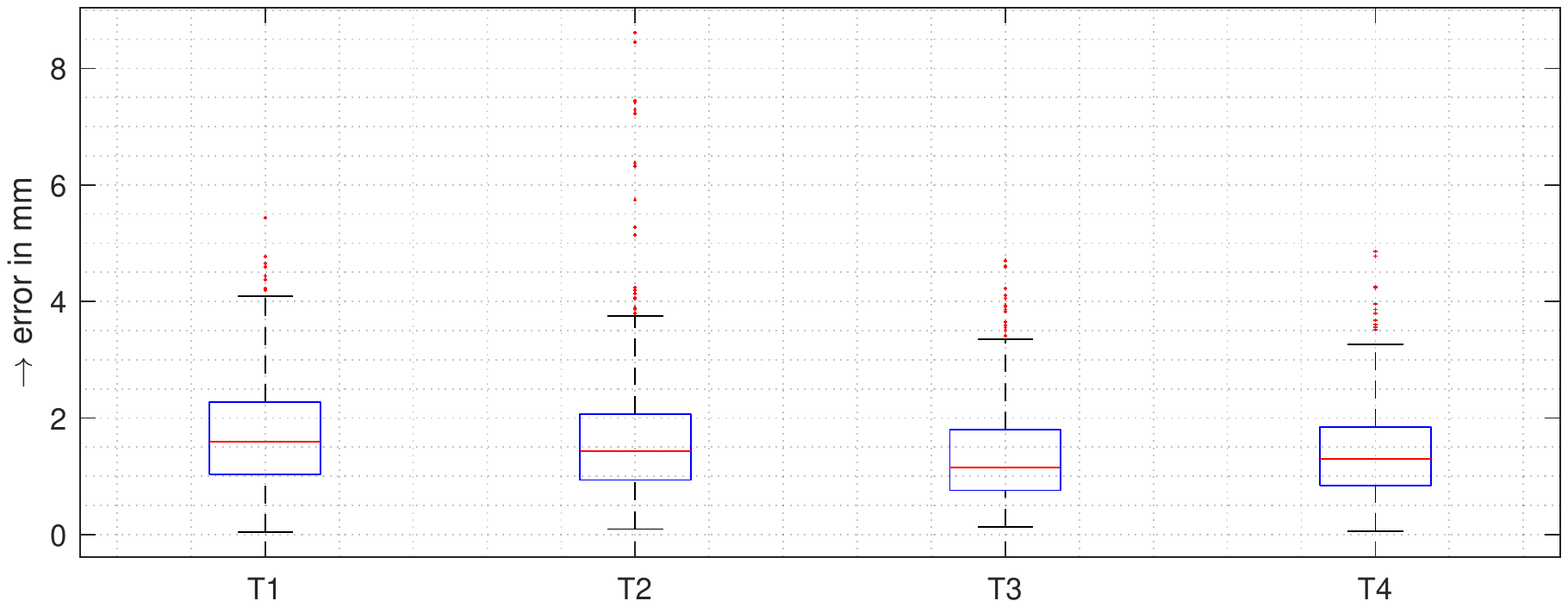}
    \caption{Teat pose error deviation for each teat (789 measurement)}
    \label{fig:evaluationDataP020201202}
\end{figure}

The first experimental question we ask concerns the accuracy in estimation of the pose of the teat tip and its repeatability.
To evaluate this, we kept the cow model in the same position and we precisely measured teat tip poses for each teat (T1-T4) to provide a ground truth.
Then we ran 789 teat position estimation cycles under varying light conditions and initial arm positions, measured the error of the pose estimation w.r.t. the ground truth obtaining the results in Figure \ref{fig:evaluationDataP020201202} and Table \ref{table:evaluationDataP020201202}.

\begin{table}[h!]
    \centering
    \begin{tabular}{c c c}
        in mm & mean & std \\
        \hline
        T1 & 1.7 & 0.9 \\
        T2 & 1.6 & 1.0 \\
        T3 & 1.4 & 0.8 \\
        T4 & 1.4 & 0.7 \\
        \hline
    \end{tabular}
    \caption{Mean and standard deviation of pose error per teat} 
    \label{table:evaluationDataP020201202}
\end{table}
\vspace{-2em}

\subsection{Pose Estimation Rate}

Both the Detectron and Matterport implementation of MaskRCNN achieve a similar inference time (on 640x480 images) of roughly 150 ms on a Tesla T4 GPU on a remote server. We are confident that further engineering of the implementation could sensibly reduce it. 
Inference performance could be trivially increased by reducing processed image resolution (at the cost of lower mask precision).
We still need to evaluate performance of the network on embedded GPU boards such as the Nvidia Jetson.
Adding an estimated latency of 50ms per submitted image over a remote connection (e.g., with 5G) even with this initial setup would yield a processing rate of 5 FPS which is sufficient for limited teat movement.

The weakest part of our current algorithm implementation is in the transformation from the 2D teat masks to the corresponding 3D points in the point cloud. The current (trivial) implementation converts each point in the contour of a mask into a 3D ray to build a (pixel-precision) frustum. This operation, calculated for each point, is currently executed sequentially resulting in an average execution time of up to 50 ms. Reducing the number of considered contour points (e.g., sampling every ten pixels) can sensibly speed up the process with limited effect on the frustum precision.

A video of the overall system prototype in operation is visible online\footnote{\url{https://www.youtube.com/watch?v=-7NiKSdA4AM}}.

\section{Conclusions and future work}\label{sec:conclusions}


This paper relates the initial work concerning the 3D sensor evaluation and teat pose estimation activities of a research project to build a next generation milking robot for the Swiss market.
The current prototype already demonstrated sub-centimeter precision in teat pose estimation (albeit on a synthetic cow). 
The presented results are preliminary and will require further engineering and validation in real environments in subsequent steps of this and following projects.

There are several directions for the extension of this work. To improve the detection rate a faster segmentation network could significantly reduce the prediction time on RGB images.

Rather than relying on a sequence of arm movements based on estimated teat poses to approach the teats, an active vision system could take into account occlusions, and learn the optimal sequence of positions to perform teat pose estimation for any shape of udder. We started working on such an ``embodied AI'' approach already in \cite{roost2020} and more work is ongoing.









\bibliographystyle{IEEEtran}
\bibliography{main}

\end{document}